\documentclass{Interspeech}

\interspeechcameraready

\usepackage{comment}
\usepackage[acronym, shortcuts, nohypertypes={acronym}]{glossaries}
\usepackage[capitalize]{cleveref}
\usepackage{tabularray}
\usepackage{amssymb,bm}
\usepackage{booktabs}
\usepackage{multirow}
\usepackage{graphicx}

\usepackage{adjustbox}
\usepackage{listings}

\newcommand{\eg}{e.\,g., }
\newcommand{\ie}{i.\,e., }

\newcommand{\et}{{et al.\ }}
\newcommand{\melt}{MELT }

\title{MELT: Towards Automated Multimodal Emotion Data Annotation by Leveraging LLM Embedded Knowledge}

\author[affiliation={1,2}]{Xin}{Jing}
\author[affiliation={1}]{Jiadong}{Wang}
\author[affiliation={1,2}]{Iosif}{Tsangko}
\author[affiliation={1,2}]{Andreas}{Triantafyllopoulos}
\author[affiliation={1,2,3}]{Björn}{W.\ Schuller}

\affiliation{CHI -- Chair of Health Informatics}{Technical University of Munich}{Germany}
\affiliation{Munich Centre for Machine Learning}{Munich}{Germany}
\affiliation{GLAM -- Group on Language, Audio, \& Music}{Imperial College London}{UK}
\email{xin.jing@tum.de}

\begin{document}

\maketitle

\begin{abstract}
    Although speech emotion recognition (SER) has advanced significantly with deep learning, annotation remains a major hurdle. Human annotation is not only costly but also subject to inconsistencies—annotators often have different preferences and may lack the necessary contextual knowledge, which can lead to varied and inaccurate labels. Meanwhile, Large Language Models (LLMs) have emerged as a scalable alternative for annotating text data. However, the potential of LLMs to perform audio data annotation without human supervision has yet to be thoroughly investigated. To address these problems, we apply GPT-4o to annotate a multimodal dataset collected from the sitcom ``Friends'', using only textual cues as inputs. By crafting structured text prompts, our methodology capitalizes on the knowledge GPT-4o has accumulated during its training, showcasing that it can generate accurate and contextually relevant annotations without direct access to multimodal inputs. Therefore, we propose \melt, a multimodal emotion dataset fully annotated by GPT-4o. We demonstrate the effectiveness of \melt by fine-tuning four self-supervised learning (SSL) backbones and assessing speech emotion recognition performance across emotion datasets. Additionally, our subjective experiments' results demonstrate a consistence performance improvement on SER. Our data can be accessed at https://github.com/KeiKinn/meltdataset.git

\end{abstract}

\section{Introduction}

Recognizing human emotion and responding accordingly is a cornerstone of human-computer interaction \cite{Triantafyllopoulos24-BDL}. The progress made by contemporary deep-learning-based emotion recognition heavily depends on the availability of well-annotated datasets \cite{Jing24-EET}. However, ensuring accurate and consistent annotation requires multiple annotators and validation, resulting in significant financial costs that limit dataset scale and diversity.

In addition, researches has demonstrated that accurately applying emotional data annotation schemes requires annotators to incorporate contextual knowledge to capture characters’ emotions effectively \cite{kosti2017emotion, etesam2024contextual}. Meanwhile, research \cite{Andalibi20-THI} indicates that individual preferences and cultural backgrounds significantly influence emotion recognition.
However, contextual understanding and individual preferences are not a primary factor in the selection of annotators. On Amazon Mechanical Turk (AMT), a widely used crowdsourcing platform for data annotation, there is a lack of qualification tests to ensure annotators' familiarity with the target samples.
With the introduction of the OpenAI’s Generative Pre-trained Transformer (GPT) models, LLMs have demonstrated the potential of performing more complex tasks with scaling \cite{Radford19-LMA, Brown20-LMA, Hurst24-GSC}. Several studies have evaluated the adaptability and broad applicability of LLMs as annotators using existing datasets \cite{Tan24-LLM, Aldeen23-CVH}.
For instance, Gilardi \et \cite{Gilardi23-COC} found that ChatGPT outperformed crowd workers by approximately 25\% points on average, with its intercoder agreement surpassing humans across all evaluated tasks. Due to the input constraints, these studies predominantly concentrate on text-based datasets. In the domain of audio, WavCaps \cite{Mei24-WAC} utilized ChatGPT to compile large-scale, high-quality audio captions, further highlighting the potential of LLMs in generating reliable annotations.
This was facilitated by the use of tags describing the audio files that comprise WavCaps; thus, ChatGPT did not introduce novel information, but rather reframed the information provided by humans.
%
Pengi \cite{Deshmukh23-PAA} introduces an Audio Language Model by reframing all audio tasks as text-generation tasks, which accepts an audio recording and a text prompt as inputs and subsequently outputs free-form text. The SECap \cite{Xu24-SSE} framework utilizes the LLaMA decoder to generate fluent and coherent captions describing emotional speech by leveraging Q-Former embeddings. 
However, these approaches not only rely on datasets with existing high-quality emotion annotations— which are limited in scale due to the high costs of collection— but also require additional audio features as input for LLM decoders to generate captions.
As a result, the potential of LLMs to automatically annotated audio datasets with captions without \emph{any} human labor, solely leveraging their contextual understanding, remains relatively underexplored, highlighting a gap in the current research landscape.

Trained on an extensive corpus of web-sourced data \cite{Hurst24-GSC, Touvron23-LOA}, frontier LLMs like GPT-4o encode knowledge regarding culturally significant content. This is especially true for widely popular media, such as the television series ``Friends''. We consider the vast corpus of internet data that the model has been trained upon as an implicit ``collective knowledge base'', reflecting the shared understanding and engagement of a broad audience.
We exploit this knowledge to re-annotate MELD -- a multimodal dataset derived from ``Friends''. 
In this work, we propose the \textbf{M}ultimodal \textbf{E}motion-Lines Dataset \textbf{L}abeled with LLM Contex\textbf{T} Knowledge (MELT) extending GPT-4o's annotation capabilities from text-only emotion data to multimodal data by leveraging its embedded knowledge.

Our accompanying experiments indicate that \melt aligns more closely with human preferences, as evidenced by subjective evaluations. Furthermore, models trained on \melt exhibit improved generalization and robustness across different speech emotion recognition (SER) datasets, highlighting LLMs' potential for tasks that extend beyond conventional text-based approaches.
This exploration demonstrates the advantages of LLMs in creating scalable and efficient annotation pipelines, while also demonstrating their ability to address complex contextual tasks.
To the best of our knowledge, this work is the first to explore the potential of GPT models as annotators for multimodal emotion data, leveraging insights from the knowledge it has assimilated in its training.

\noindent
The summary of our contributions is as follows:
\begin{itemize}
    \item We propose a context-aware automatic annotation method using GPT-4o, with consistent emotional tone across samples, resulting in the MELT, which outperforms MELD, the human-labeled counterpart.
    \item We propose a prompting framework with cross-validation and Chain-of-thoughts (CoT) reasoning for multimodal emotion annotation.
    \item A significant reduction in costs ($\le$ \$10) to achieve high-quality annotations compared to traditional methods.
\end{itemize}

The rest of this paper is organized as follows:
\cref{sec:propsed} presents our methodology, focusing on prompting framework. \cref{sec:data} provides a detailed analysis of \melt. The objective and subjective experiments are described in \cref{sec:exp}.
Results and analysis are presented and discussed in \cref{sec:res}. Finally, we conclude the paper and discuss the future work in \cref{sec:con}.

\section{Methodology}
\label{sec:propsed}
\subsection{Data Preparation}
\label{ssec:dp}
Multimodal EmotionLines Dataset (MELD) \cite{Pria19-MAMM}, built from the TV series ``Friends'', comprises 1,433 dialogues and 13,708 utterances. Each utterance is annotated with one of seven categories (Joy, Sadness, Fear, Anger, Surprise, Disgust, Neutral) based on a majority vote among three annotators.
\melt is derived from MELD using the following steps: Firstly, utterances shorter than one second were excluded, as classifying short speech remains a significant challenge in SER \cite{Pell11-OTT, Kumawat21-ATA}. Subsequently, we excluded characters whose names do not provide enough context for GPT-4o to maintain consistency.
MELD includes 260 unique characters in the training set and 100 in the test set, with some overlap. Certain characters, such as ``\textit{1st Customer}'' and ``\textit{Receptionist},'' lack clear identifiers like names or gender in the textual modality, which conflicted with the prompt design guidelines in \cref{ssec:prompt}. 

\subsection{GPT Model Selection}

We utilize the OpenAI API\footnote{https://openai.com/api} to access the `gpt-4o-2024-08-06' model with a temperature of 1.0 for speech emotion annotation. For simplicity, `GPT-4o' is used throughout the following sections. GPT-4o, with its October 2023 knowledge base cutoff\footnote{https://platform.openai.com/docs/models}, integrates updated data, reducing reliance on fine-tuning or retrieval-augmented generation (RAG) methods \cite{Hurst24-GSC}.
 
\subsection{Prompt Engineering}
\label{ssec:prompt}
Effective prompt engineering significantly impacts the performance of LLMs \cite{Amin24-AWE}.  To design a prompt that optimize performance while ensuring stability and reproducibility, we adhere to the following principles:
\begin{itemize}
    \item \textbf{Clear, Contextual, and Specific}: Include as much relevant context as possible while avoiding ambiguity in instructions to enhance the model's understanding of the task.  
    \item \textbf{Chain of Thought (CoT) Prompting}: Break tasks into distinct, logical steps to guide the model’s reasoning process.  
    \item \textbf{Cross-Validation}: Incorporate requests for known or easily verifiable information to reduce the likelihood of generating incorrect or unrelated content.  
    \item \textbf{Guide Output with Prefilling Responses}: Structure prompts (\eg JSON or XML) to direct the model's output to ensure consistency and ease of post-processing.   
\end{itemize}  
\begin{lstlisting}
Given the following line of dialogue from a Friends character, the format will be:
"[speaker] at s[season]e[episode] said: [utterance]"

Please describe how the character's voice might sound. Include details about:
- the emotion expressed,
- the loudness,
- the pitch,
- the rhythm speed,
- the overall emotional impact of the voice.

Format your response:
- Provide the character's name and a brief context.
- The emotion label must be selected from the following list: [Anger, Disgust, Sadness, Joy, Neutral, Surprise, Fear].

Format the response in the following JSON structure:
{
    "character": "[Character's name]",
    "context": "[Simple context of the situation]",
    "elements": {
        "emotion": "[Primary emotion]",
        "loudness": "[Loudness of voice]",
        "pitch": "[Pitch of voice]",
        "rhythm_speed": "[Speed of voice's rhythm]",
        "emotional_impact": "[Emotional impact created by the voice]"
    }
}
\end{lstlisting}

\section{The \melt Corpus}
\label{sec:data}
Following \cref{ssec:dp}, \melt retains 42 unique speakers across both the training and test sets, ensuring consistency and fair evaluation. As summarized in \cref{tab:overview}, approximately 70\% of the original utterances in MELD have been preserved.

\begin{table}[h]
\centering
\caption{Summary statistics of \melt and MELD for training and test sets. \#Utt, \#Spk, \#Avg.\ Sec.\ represents utterance number, speaker number, and average seconds, respectively.}
    \begin{tabular}{@{}lcccc@{}}
    \toprule
    \textbf{Dataset} & \textbf{Split} & \textbf{\#Utt} & \textbf{\#Spk } & \textbf{\#Avg. Sec.}\\ 
    \midrule
    \multirow{2}{*}{\melt} & Train & 7024 & 42 &3.68 \\ 
                            & Test  & 1797 & 42 &3.82 \\ 
    \midrule
    \multirow{2}{*}{MELD} & Train  & 9989 & 260 &3.14 \\ 
                          & Test   & 2610 & 100 &3.29 \\ 
    \bottomrule
    \end{tabular}
\label{tab:overview}
\end{table}

\begin{table}[ht]
\centering
\caption{Distribution of emotions in MELT and filtered MELD for training and test splits. label $\Delta$\% is the label change rate}
\begin{tabular}{@{}lrrrrrr@{}}
\toprule
\multirow{2}{*}{\textbf{Emo.}} & \multicolumn{2}{c}{\textbf{Train Set}} & \phantom{a} & \multicolumn{2}{c}{\textbf{Test Set}} \\ 
\cmidrule{2-3} \cmidrule{5-6}
                 & \textbf{\melt} & \textbf{MELD} && \textbf{\melt} & \textbf{MELD} \\ 
\midrule
anger              & 464           & 852           && 127           & 270           \\ 
disgust              & 473           & 215           && 124           & 59            \\ 
fear              & 418           & 208           && 103           & 42            \\ 
joy              & 1578          & 1244          && 389           & 262           \\ 
neutral              & 2519          & 3172          && 660           & 827           \\ 
sadness              & 363           & 570           && 90            & 170           \\ 
surprise              & 1209          & 763           && 304           & 167           \\ 
\midrule
label $\Delta$\% & \multicolumn{2}{c}{\textbf{46.43}} & \phantom{a} & \multicolumn{2}{c}{\textbf{47.52}}\\
\midrule
\textbf{Total}   & 7024          & 7024          && 1797          & 1797          \\ 
\bottomrule
\end{tabular}
\vspace{-5mm}
\label{tab:distribution}
\end{table}

\begin{figure}[ht]
    \centering
    \includegraphics[width=0.48\textwidth]{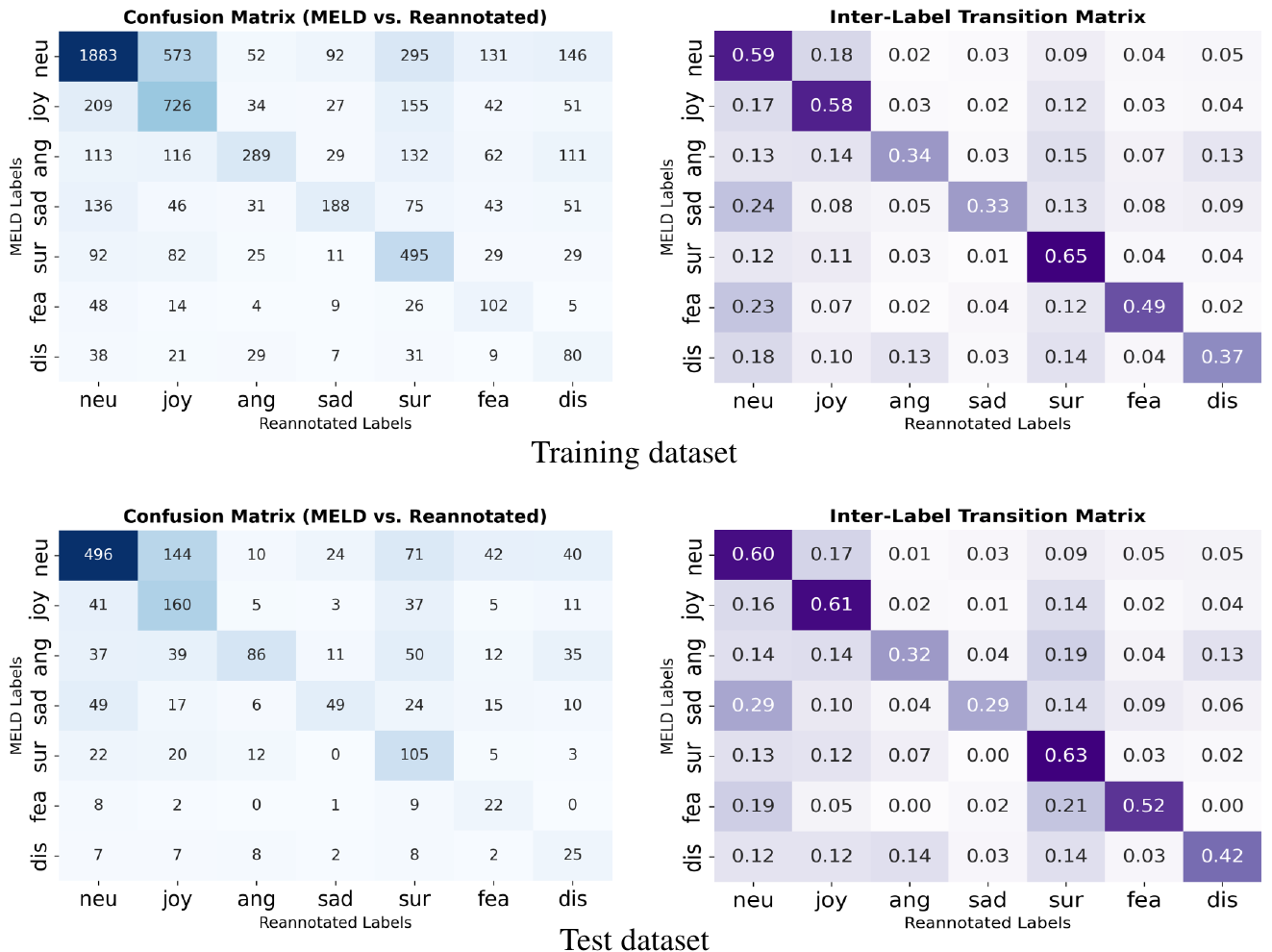}
    \caption{Confusion matrix and inter-label transition matrix of the training and test dataset}
    \vspace{-5mm}
    \label{fig:train_data}
\end{figure}

\cref{tab:distribution} compares the label distribution and statistical characteristics of the \melt and MELD datasets. Both datasets are dominated by the `neutral' emotion, followed by `joy', while `anger', `sadness', and `fear' are underrepresented. Notably, \melt shows a more balanced distribution, with training and test sets maintaining an approximate 4:1 ratio across labels.
As shown in \cref{fig:train_data}, nearly half of MELD's utterances align with GPT-4o's annotations, and the overall label proportions between the two datasets are similar for most emotions. However, `sad' and `anger' exhibit significant reannotation, with most samples being relabeled as `neutral' and `surprise', respectively. This likely reflects overlapping features or ambiguous original annotations.
While labels such as `neutral' and `joy' remain consistent, emotions like `fear' and `disgust' are more frequently reannotated, potentially due to inherent ambiguity. The percentage of label changes is comparable in both datasets (46.43\% in training and 47.52\% in testing), with slightly higher changes in the test set. 

\begin{table*}[ht]
\centering
\vspace{-5mm}
\caption{Speech Emotion Recognition Performance of the Evaluation system on four Emotion Recognition Datasets. The `$\#$cl' indicates the emotion category of the dataset. The best scores for each metric are highlighted in \textbf{bold}.}
\resizebox{\linewidth}{!}{
\begin{tabular}{@{}llcccccccccccc@{}}
\toprule
\textbf{Backbone} & \textbf{Annotation} & 
\multicolumn{3}{c}{\textbf{IEMOCAP (4\,cl)}} &
\multicolumn{3}{c}{\textbf{TESS (7\,cl)}} &
\multicolumn{3}{c}{\textbf{RAVDESS (7\,cl)}} &
\multicolumn{3}{c}{\textbf{CREMA-D (6\,cl)}} \\ 
\cmidrule(lr){3-5} \cmidrule(lr){6-8} \cmidrule(lr){9-11} \cmidrule(lr){12-14}
 &  & \textbf{UAR} & \textbf{ACC} & \textbf{F1} & \textbf{UAR} & \textbf{ACC} & \textbf{F1} & \textbf{UAR} & \textbf{ACC} & \textbf{F1} & \textbf{UAR} & \textbf{ACC} & \textbf{F1} \\
\midrule

wav2vec 2.0 Base& \melt   & \textbf{0.3805} & \textbf{0.4262} & \textbf{0.3610} & \textbf{0.3407} & \textbf{0.3407} & \textbf{0.2567}        & \textbf{0.2835} & \textbf{0.2284} & \textbf{0.1799} & \textbf{0.2789} & \textbf{0.2712} & \textbf{0.2269} \\
        & MELD & 0.3565 & 0.4014 & 0.3152 & 0.2857 & 0.2857 & 0.1614         & 0.2284 & 0.1691 & 0.0877 & 0.2752 & 0.2607 & 0.2034 \\
\midrule

wav2vec 2.0 Aud   & \melt   & \textbf{0.4991} & \textbf{0.4899} & \textbf{0.4870}         & \textbf{0.2946} & \textbf{0.2946} & 0.1926         & \textbf{0.2753} & \textbf{0.2200} & \textbf{0.1534} & \textbf{0.2328} & \textbf{0.2194} & \textbf{0.1748} \\
        & MELD & 0.4347 & 0.4523 & 0.4181 & 0.2943 & 0.2943 & \textbf{0.2063} & 0.2292 & 0.1699 & 0.0849 & 0.2017 & 0.1819 & 0.0986\\
\midrule

Hubert Base& \melt   & \textbf{0.3675} & \textbf{0.4079} & \textbf{0.3415} & \textbf{0.3107} & \textbf{0.3107} & \textbf{0.1879} & \textbf{0.3013} & \textbf{0.2516} & \textbf{0.2047} & \textbf{0.2419} & \textbf{0.2327} & \textbf{0.1697} \\
       & MELD & 0.3296 & 0.3771 & 0.2930         & 0.2511 & 0.2511 & 0.1730         & 0.2433 & 0.1851 & 0.1577 & 0.2380 & 0.2202 & 0.1526 \\
\midrule

WavLM Base+  & \melt   & \textbf{0.4568} & \textbf{0.4887} & \textbf{0.4575}         & 0.3114 & 0.3114 & 0.2372         & \textbf{0.3244} & \textbf{0.2724} & \textbf{0.2174} & \textbf{0.3135} & \textbf{0.3007} & \textbf{0.2480} \\
        & MELD & 0.4188 & 0.4570 & 0.4102 & 0.3114 & 0.3114 & \textbf{0.2405} & 0.2827 & 0.2276 & 0.1553 & 0.3069 & 0.2912 & 0.2018 \\
\bottomrule

\end{tabular}}
\vspace{-5mm}
\label{tab:main}
\end{table*}

\begin{figure}[ht]
    \centering
    \includegraphics[width=0.42\textwidth]{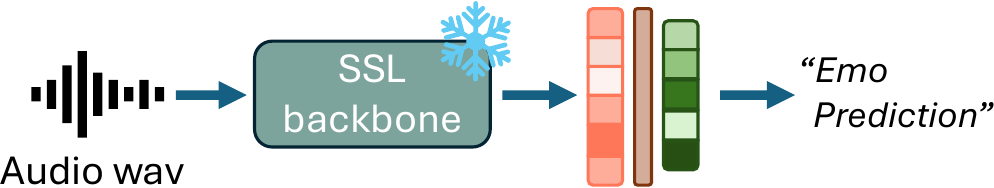}
    \caption{Speech emotion recognition evaluation system}
    \label{fig:eval_sys}
\end{figure}

\section{Experiments}
\label{sec:exp}
\subsection{Subjective Experiment}
\label{ssec:sub_expr}
To assess the emotion annotation quality, we invited 20 participants, comprising 11 males and 9 females to conduct a Mean Opinion Score~(MOS) experiment. Participants were instructed to watch the video clips and were presented with two options—\melt and MELD annotations, without knowing their sources —from which they were asked to select the description they deemed more appropriate.

\subsection{Objective Experiments}
\label{sec:obj_expr}
\subsubsection{Objective Evaluation system}

The SER evaluation system in \cref{fig:eval_sys} consists of two main components: 1) a pretrained Self-supervised learning (SSL) backbone, which is initialized by the pretrained weight on \texttt{huggingface.co}\footnote{https://huggingface.co/\{backbone\_model\}} and 2) a classification module. 

We applied four pretrained SSL weights:  `\texttt{facebook/wav2vec2-base-960h\cite{Baevski20-WAF}}' (wav2vec 2.0 base) , `\texttt{audeering/wav2vec2-large-\\robust-12-ft-emotion-msp-dim \cite{WAGNER23-DOT}}' (wav2vec 2.0 Aud), `\texttt{facebook/hubert-base-ls960 \cite{Hsu21-HSS}}' (Hubert Base), and `\texttt{microsoft/wavlm-base-plus \cite{Chen22-WLS}}' (WavLM Base+).
The classification module consists of two fully connected layers with a ReLU activation function between them. The classification head is dynamically adjusted to match the number of emotion categories in the test dataset. We apply the weighted sum of hidden states as the input of this classification module.

\subsubsection{Datasets}

\label{ssec:dataset}

\textbf{Training Datasets}: Both \melt and MELD are employed as training datasets in our evaluation system. To ensure a fair comparison, we impose two conditions on data filtering: 1) only emotion categories present in the test set are retained by filtering out others, and 2) training data is restricted to audio samples that are common to both datasets. 

\noindent
\textbf{Out-of-domain Datasets:}
We additionally perform cross-corpus testing on the following datasets to test the generalizability of the trained models.

\noindent
\textbf{IEMOCAP \cite{Busso08-IIED}:} 
We follow the convention in the literature to use a subset of the original labels by merging `happy' and `excited', 
resulting total of 5,531 utterances, with four labels (\textbf{1,103 angry, 1,636 happy, 1,708 neutral, and 1,084 sad}). 

\noindent
\textbf{TESS \cite{Pichora20-TESS}:} The set consists of 2,800 clips representing seven emotions: \textbf{anger, disgust, fear, happiness, pleasant surprise, sadness}, and \textbf{neutral}. To align with \melt, we set the data labeled `pleasant surprise' to `surprise'. 

\noindent
\textbf{RAVDESS \cite{Livingstone18-TRAD}:} Its speech portion includes 1,440 utterances across 8 emotion categories: \textbf{neutral, calm, happy, sad, angry, fearful, surprise, disgust}. During the evaluation, we excluded the `calm' category, resulting 1,248 utterances in test set.

\noindent
\textbf{CREMA-D \cite{Cao14-CCEM}:} It contains 7,442 clips recorded, featuring both facial and vocal expressions across six basic emotional states: \textbf{happy, sad, angry, fearful, disgust}, and \textbf{neutral}.

\subsection{Experiment Settings}

The training process employed the Adam optimizer with a batch size of 32, a learning rate of 0.001, and a dropout rate of 0.2 for 100 epochs. Audio data are resampled to 16\,kHz and randomly cropped or padded to 5 seconds. During testing, batch size was set to 1, and no length adjustment was required. All experiments were conducted in a Python 3.10.8 and PyTorch 2.4.1 environment on a single Nvidia RTX 3090 GPU.

\section{Results and Analysis}
\label{sec:res}
\subsection{Performance}
\begin{figure}
    \centering
    \vspace{3mm}
    \includegraphics[width=0.45\textwidth]{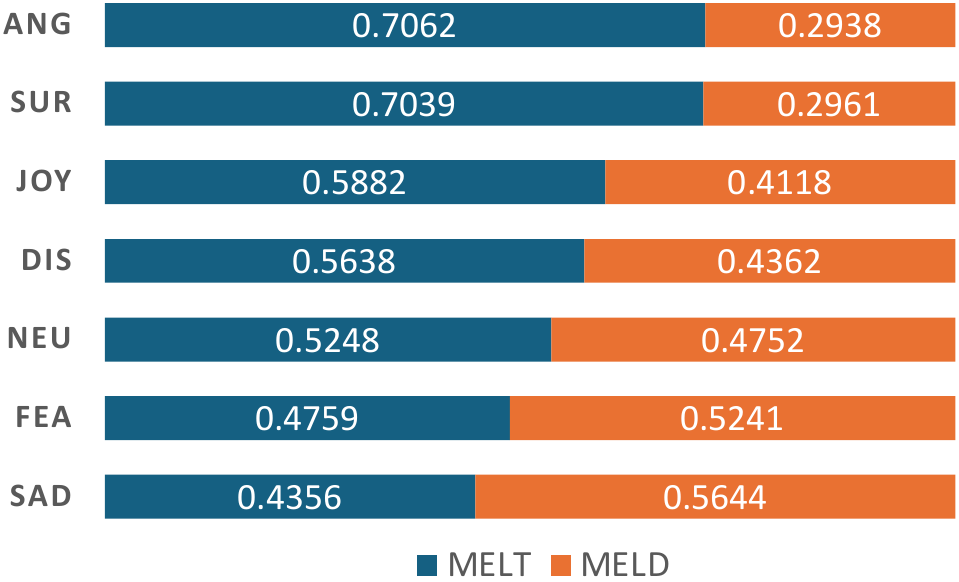}
    \caption{MOS results on every emotion categories.}

    \label{fig:mos}
\end{figure}

The overall MOS result is shown in \cref{fig:mos}, participants overall demonstrated a preference for \melt annotations. 
The high agreement ($> 70\%$) for `anger' and `surprise' indicates that GPT-4o effectively integrates internet-sourced knowledge to capture the diversity in expressions.
However, an opposite trend is observed for `fear' and `sadness', as they mostly transition to `neutral', as shown in \cref{fig:train_data}. 
These reannotated audio clips often feature lower arousal and subtler emotional contexts. The comparable preferences suggest possible biases stemming from GPT-4o's reliance on internet-derived emotional knowledge.

Classification results are summarized in \cref{tab:main}. Commonly used evaluation metrics, unweighted accuracy recall (UAR), accuracy (ACC), and F1 score, are applied to evaluate the performance of the SER task.  Our first observation is that the in-domain results trained on \melt generally outperform the original MELD, showcasing how our annotation improves upon the general SSL-based benchmark architecture \cite{Wu24-EAI}.

For the TESS and RAVDESS, where \melt shares the same emotion classes, there existed two different patterns in the experimental results. For RAVDESS, all models demonstrated considerable
and consistent performance improvements. Meanwhile, TESS showed a mixed outcome: while wav2vec 2.0 Base and HuBERT Base achieved substantial UAR performance gains (19.25\% and 23.74\%, respectively), the remaining two models exhibited minimal improvements in UAR and ACC, with a slight decline in F1 scores compared to results on the MELD dataset. 
For IEMOCAP and CREMA-D, which only partially overlap with \melt, the test performance varies across different model backbones. wav2vec 2.0 Aud achieves the most substantial performance improvement on both datasets, while wav2vec 2.0 Base shows relatively modest gains. This highlights the importance of pretraining in enhancing model performance, particularly when addressing datasets with limited or partial alignment to the training annotations.

\subsection{Audio characteristic}
To further analyze the annotation performance of the GPT-4o on audio characteristics, we adapted the query generation pipeline proposed by ParaCLAP \cite{Jing24-PTA} to categorize descriptions and extract the extended Geneva Minimalistic Acoustic Parameter Set (eGeMAPS)~ \cite{Eyben15-TGM} using openSMILE~ \cite{Eyben10-OTM}. 
The pitch and loudness attributes are binned according to their distribution (bottom 30\,\%, middle 40\,\%, top 30\,\%) and pseudo-captions are mapped accordingly (\ie low/mid/high). 

\begin{table}[h!]
    \centering
    \vspace{-2mm}
    \caption{Performance metrics for pitch and loudness.}
    \begin{tabular}{llccc}
        \toprule
        Char. &       Data Split         & \textbf{UAR} & \textbf{ACC} & \textbf{F1} \\ \midrule
        \multirow{2}{*}{\textbf{Pitch}}   & Train & 0.4761 & 0.5213 & 0.4376 \\ 
                                          & Test  & 0.4880 & 0.5493 & 0.4534 \\ \midrule
        \multirow{2}{*}{\textbf{Loud.}} & Train & 0.4103 & 0.4546 & 0.4005 \\ 
                                           & Test  & 0.3904 & 0.4374 & 0.3824 \\ \bottomrule
    \end{tabular}
    \vspace{-2mm}
    \label{tab:metrics}
\end{table}

Across both the training and test sets, all metrics consistently surpass random guessing, demonstrating that GPT-4o effectively captures relevant characteristics from its embedded knowledge. Notably, the performance of pitch exceeds that of loudness, likely reflecting human preferences in voice descriptions \cite{Lavan23-HDW}, which may contribute to biases in the output. These outcomes, while still falling short of state-of-the-art results, showcase the potential of LLMs with further refinement for tasks traditionally dominated by handcrafted features.

\section{Conclusion}
\label{sec:con}
In this work, we introduce \melt, a multimodal emotion dataset fully annotated by GPT-4o using a context-aware automatic annotation method. 
To achieve this, we developed a prompting strategy incorporating cross-validation and CoT reasoning to ensure consistent and accurate annotations. The MOS and classification results highlights that \melt provides better generalization and robustness, demonstrating the feasibility of adapting LLMs for multimodal tasks and sets the stage for future innovations in multimodal annotation frameworks. 
Despite the promising results, the reliance on GPT-4o raises concerns about model-specific biases and hallucinations, which can affect annotation quality. Future work will aim to investigate hybrid annotation pipelines that combine LLMs with human-in-the-loop methods to improve scalability and reliability. 
\section{Acknowledgment}
\ifinterspeechfinal
     Björn W.\ Schuller is also with the Munich Data Science Institute and the Konrad Zuse School of Excellence in Reliable AI, both in Munich, Germany. We are grateful to the China Scholarship Council~(CSC), Grant \#\,202006290013 to support this work. We would also like to thank Prof.\ Xinzhou Xu and everyone involved in the MOS experiments for their invaluable contributions.
\else
     The authors would like to thank ISCA and the organising committees of past Interspeech conferences for their help and for kindly providing the previous version of this template.
\fi

\bibliographystyle{IEEEtran}
\bibliography{mybib}

\begin{thebibliography}{10}
\providecommand{\url}[1]{#1}
\csname url@samestyle\endcsname
\providecommand{\newblock}{\relax}
\providecommand{\bibinfo}[2]{#2}
\providecommand{\BIBentrySTDinterwordspacing}{\spaceskip=0pt\relax}
\providecommand{\BIBentryALTinterwordstretchfactor}{4}
\providecommand{\BIBentryALTinterwordspacing}{\spaceskip=\fontdimen2\font plus
\BIBentryALTinterwordstretchfactor\fontdimen3\font minus \fontdimen4\font\relax}
\providecommand{\BIBforeignlanguage}[2]{{%
\expandafter\ifx\csname l@#1\endcsname\relax
\typeout{** WARNING: IEEEtran.bst: No hyphenation pattern has been}%
\typeout{** loaded for the language `#1'. Using the pattern for}%
\typeout{** the default language instead.}%
\else
\language=\csname l@#1\endcsname
\fi
#2}}
\providecommand{\BIBdecl}{\relax}
\BIBdecl

\bibitem{Triantafyllopoulos24-BDL}
A.~Triantafyllopoulos, L.~Christ, A.~Gebhard, X.~Jing, A.~Kathan, M.~Milling, I.~Tsangko, S.~Amiriparian, and B.~W. Schuller, ``Beyond deep learning: Charting the next frontiers of affective computing,'' \emph{Intelligent Computing}, vol.~3, p. 0089, 2024.

\bibitem{Jing24-EET}
X.~Jing, K.~Zhou, A.~Triantafyllopoulos, and B.~W. Schuller, ``Enhancing emotional text-to-speech controllability with natural language guidance through contrastive learning and diffusion models,'' in \emph{Proceedings of International Conference on Acoustics, Speech and Signal Processing (ICASSP)}, Hyderabad,India, 2025, pp. 1--5.

\bibitem{kosti2017emotion}
R.~Kosti, J.~M. Alvarez, A.~Recasens, and A.~Lapedriza, ``Emotion recognition in context,'' in \emph{Proceedings of the IEEE conference on computer vision and pattern recognition}, 2017, pp. 1667--1675.

\bibitem{etesam2024contextual}
Y.~Etesam, {\"O}.~N. Yal{\c{c}}{\i}n, C.~Zhang, and A.~Lim, ``Contextual emotion recognition using large vision language models,'' \emph{arXiv preprint arXiv:2405.08992}, 2024.

\bibitem{Andalibi20-THI}
N.~Andalibi and J.~Buss, ``The human in emotion recognition on social media: Attitudes, outcomes, risks,'' in \emph{Proceedings of the 2020 CHI Conference on Human Factors in Computing Systems}, New York, NY, USA, 2020, p. 1–16.

\bibitem{Radford19-LMA}
A.~Radford, J.~Wu, R.~Child, D.~Luan, D.~Amodei, I.~Sutskever \emph{et~al.}, ``Language models are unsupervised multitask learners,'' \emph{OpenAI blog}, vol.~1, no.~8, p.~9, 2019.

\bibitem{Brown20-LMA}
T.~B. Brown, ``Language models are few-shot learners,'' \emph{arXiv preprint arXiv:2005.14165}, 2020.

\bibitem{Hurst24-GSC}
A.~Hurst, A.~Lerer, A.~P. Goucher, A.~Perelman, A.~Ramesh, A.~Clark, A.~Ostrow, A.~Welihinda, A.~Hayes, A.~Radford \emph{et~al.}, ``Gpt-4o system card,'' \emph{arXiv preprint arXiv:2410.21276}, 2024.

\bibitem{Tan24-LLM}
Z.~Tan, D.~Li, S.~Wang, A.~Beigi, B.~Jiang, A.~Bhattacharjee, M.~Karami, J.~Li, L.~Cheng, and H.~Liu, ``Large language models for data annotation: A survey,'' \emph{arXiv preprint arXiv:2402.13446}, 2024.

\bibitem{Aldeen23-CVH}
M.~Aldeen, J.~Luo, A.~Lian, V.~Zheng, A.~Hong, P.~Yetukuri, and L.~Cheng, ``Chatgpt vs. human annotators: A comprehensive analysis of chatgpt for text annotation,'' in \emph{Proceedings of International Conference on Machine Learning and Applications (ICMLA)}, Florida, USA, 2023, pp. 602--609.

\bibitem{Gilardi23-COC}
F.~Gilardi, M.~Alizadeh, and M.~Kubli, ``Chatgpt outperforms crowd workers for text-annotation tasks,'' \emph{Proceedings of the National Academy of Sciences}, vol. 120, no.~30, 2023.

\bibitem{Mei24-WAC}
X.~Mei, C.~Meng, H.~Liu, Q.~Kong, T.~Ko, C.~Zhao, M.~D. Plumbley, Y.~Zou, and W.~Wang, ``Wavcaps: A chatgpt-assisted weakly-labelled audio captioning dataset for audio-language multimodal research,'' \emph{IEEE/ACM Transactions on Audio, Speech, and Language Processing}, vol.~32, pp. 3339--3354, 2024.

\bibitem{Deshmukh23-PAA}
S.~Deshmukh, B.~Elizalde, R.~Singh, and H.~Wang, ``Pengi: An audio language model for audio tasks,'' \emph{Advances in Neural Information Processing Systems}, vol.~36, pp. 18\,090--18\,108, 2023.

\bibitem{Xu24-SSE}
Y.~Xu, H.~Chen, J.~Yu, Q.~Huang, Z.~Wu, S.-X. Zhang, G.~Li, Y.~Luo, and R.~Gu, ``Secap: Speech emotion captioning with large language model,'' in \emph{Proceedings of the AAAI Conference on Artificial Intelligence}, vol.~38, 2024, pp. 19\,323--19\,331.

\bibitem{Touvron23-LOA}
H.~Touvron, T.~Lavril, G.~Izacard, X.~Martinet, M.-A. Lachaux, T.~Lacroix, B.~Rozi{\`e}re, N.~Goyal, E.~Hambro, F.~Azhar \emph{et~al.}, ``Llama: Open and efficient foundation language models,'' \emph{arXiv preprint arXiv:2302.13971}, 2023.

\bibitem{Pria19-MAMM}
S.~Poria, D.~Hazarika, N.~Majumder, G.~Naik, E.~Cambria, and R.~Mihalcea, ``Meld: A multimodal multi-party dataset for emotion recognition in conversations,'' in \emph{Proceedings of Annual Meeting of the Association for Computational Linguistics}, Florence, Italy, 2019, pp. 527--536.

\bibitem{Pell11-OTT}
M.~D. Pell and S.~A. Kotz, ``On the time course of vocal emotion recognition,'' \emph{PlOS one}, vol.~6, no.~11, p. e27256, 2011.

\bibitem{Kumawat21-ATA}
P.~Kumawat and A.~Routray, ``Applying tdnn architectures for analyzing duration dependencies on speech emotion recognition.'' in \emph{Interspeech}, 2021, pp. 3410--3414.

\bibitem{Amin24-AWE}
M.~M. Amin, R.~Mao, E.~Cambria, and B.~W. Schuller, ``A wide evaluation of chatgpt on affective computing tasks,'' \emph{IEEE Transactions on Affective Computing}, vol.~15, no.~4, pp. 2204--2212, 2024.

\bibitem{Baevski20-WAF}
A.~Baevski, Y.~Zhou, A.~Mohamed, and M.~Auli, ``wav2vec 2.0: A framework for self-supervised learning of speech representations,'' \emph{Advances in neural information processing systems}, vol.~33, pp. 12\,449--12\,460, 2020.

\bibitem{WAGNER23-DOT}
J.~Wagner, A.~Triantafyllopoulos, H.~Wierstorf, M.~Schmitt, F.~Burkhardt, F.~Eyben, and B.~W. Schuller, ``Dawn of the transformer era in speech emotion recognition: closing the valence gap,'' \emph{IEEE Transactions on Pattern Analysis and Machine Intelligence}, vol.~45, no.~9, pp. 10\,745--10\,759, 2023.

\bibitem{Hsu21-HSS}
W.-N. Hsu, B.~Bolte, Y.-H.~H. Tsai, K.~Lakhotia, R.~Salakhutdinov, and A.~Mohamed, ``Hubert: Self-supervised speech representation learning by masked prediction of hidden units,'' \emph{IEEE/ACM transactions on audio, speech, and language processing}, vol.~29, pp. 3451--3460, 2021.

\bibitem{Chen22-WLS}
S.~Chen, C.~Wang, Z.~Chen, Y.~Wu, S.~Liu, Z.~Chen, J.~Li, N.~Kanda, T.~Yoshioka, X.~Xiao \emph{et~al.}, ``Wavlm: Large-scale self-supervised pre-training for full stack speech processing,'' \emph{IEEE Journal of Selected Topics in Signal Processing}, vol.~16, no.~6, pp. 1505--1518, 2022.

\bibitem{Busso08-IIED}
C.~Busso, M.~Bulut, C.-C. Lee, A.~Kazemzadeh, E.~Mower, S.~Kim, J.~N. Chang, S.~Lee, and S.~S. Narayanan, ``Iemocap: Interactive emotional dyadic motion capture database,'' \emph{Language resources and evaluation}, vol.~42, pp. 335--359, 2008.

\bibitem{Pichora20-TESS}
\BIBentryALTinterwordspacing
M.~K. Pichora-Fuller and K.~Dupuis, ``Toronto emotional speech set (tess).''\hskip 1em plus 0.5em minus 0.4em\relax Borealis, 2020. [Online]. Available: \url{https://doi.org/10.5683/SP2/E8H2MF}
\BIBentrySTDinterwordspacing

\bibitem{Livingstone18-TRAD}
S.~R. Livingstone and F.~A. Russo, ``The ryerson audio-visual database of emotional speech and song (ravdess): A dynamic, multimodal set of facial and vocal expressions in north american english,'' \emph{PloS one}, vol.~13, no.~5, 2018.

\bibitem{Cao14-CCEM}
H.~Cao, D.~G. Cooper, M.~K. Keutmann, R.~C. Gur, A.~Nenkova, and R.~Verma, ``Crema-d: Crowd-sourced emotional multimodal actors dataset,'' \emph{IEEE Transactions on Affective Computing}, vol.~5, no.~4, pp. 377--390, 2014.

\bibitem{Wu24-EAI}
H.~Wu, H.-C. Chou, K.-W. Chang, L.~Goncalves, J.~Du, J.-S.~R. Jang, C.-C. Lee, and H.-Y. Lee, ``Emo-superb: An in-depth look at speech emotion recognition,'' \emph{arXiv preprint arXiv:2402.13018}, 2024.

\bibitem{Jing24-PTA}
X.~Jing, A.~Triantafyllopoulos, and B.~Schuller, ``Paraclap--towards a general language-audio model for computational paralinguistic tasks,'' in \emph{Proceedings of International Speech Communication Association (INTERSPEECH)}, Kos Island, Greece, September 2024.

\bibitem{Eyben15-TGM}
F.~Eyben, K.~R. Scherer, B.~W. Schuller, J.~Sundberg, E.~Andr{\'e}, C.~Busso, L.~Y. Devillers, J.~Epps, P.~Laukka, S.~S. Narayanan \emph{et~al.}, ``The geneva minimalistic acoustic parameter set (gemaps) for voice research and affective computing,'' \emph{IEEE transactions on affective computing}, vol.~7, no.~2, pp. 190--202, 2015.

\bibitem{Eyben10-OTM}
F.~Eyben, M.~W{\"o}llmer, and B.~Schuller, ``{openSMILE}: the {M}unich versatile and fast open-source audio feature extractor,'' in \emph{Proc. the International Conference on Multimedia}, 2010, pp. 1459--1462.

\bibitem{Lavan23-HDW}
N.~Lavan, ``How do we describe other people from voices and faces?'' \emph{Cognition}, vol. 230, p. 105253, 2023.

\end{thebibliography}
\end{document}